\documentclass[10pt,twocolumn,letterpaper]{article}

\usepackage[pagenumbers]{cvpr} 

\usepackage{bbding}
\usepackage{bbding}
\usepackage{rotating}
\usepackage{times}
\usepackage{epsfig}
\usepackage{graphicx}
\usepackage{amsmath}
\usepackage{amssymb}
\usepackage{comment}
\usepackage{threeparttable}

\usepackage[utf8]{inputenc} 
\usepackage[T1]{fontenc}    
\usepackage{url}            
\usepackage{booktabs}       
\usepackage{amsfonts}       
\usepackage{nicefrac}       
\usepackage{microtype}      
\usepackage{xcolor}         
\usepackage{colortbl}

\usepackage{multirow}
\usepackage{enumitem}
\usepackage{makecell}
\usepackage{pifont}

\usepackage{arydshln}
\usepackage{bbm}
\usepackage[export]{adjustbox}
\usepackage{wrapfig}

\usepackage{titletoc}

\usepackage{pifont} 

\definecolor{LightGray}{rgb}{0.9,0.9,0.9}
\definecolor{DarkRed}{rgb}{0.75,0,0}
\definecolor{DarkBlue}{rgb}{0,0,0.55}
\definecolor{DarkGreen}{rgb}{0.43, 0.68, 0.28}

\definecolor{xishen}{rgb}{0.858, 0.188, 0.478}

\definecolor{DarkBlue}{rgb}{0.0, 0.5, 0.8}

\newcommand{\llf}[1]{\textcolor{black}{ #1}}

\definecolor{cvprblue}{rgb}{0.21,0.49,0.74}
\usepackage[pagebackref,breaklinks,colorlinks,allcolors=cvprblue]{hyperref}

\def\paperID{6901} 
\def\confName{CVPR}
\def\confYear{2026}

\title{A Closer Look at Cross-Domain Few-Shot Object Detection: Fine-Tuning Matters and Parallel Decoder Helps}

\author{
    Xuanlong Yu$^1$ \quad 
    Youyang Sha$^1$ \quad 
    Longfei Liu$^1$ \quad 
    Xi Shen$^{1~  \text{\Envelope}}$ \quad 
    Di Yang$^{2~  \text{\Envelope}}$ \\[6pt] 
    $^1$Intellindust AI Lab \qquad
    $^2$Suzhou Institute for Advanced Research, USTC
}

\begin{document}
\maketitle
\begin{abstract}
Few-shot object detection (FSOD) is challenging due to unstable optimization and limited generalization arising from the scarcity of training samples. To address these issues, we propose a hybrid ensemble decoder that enhances generalization during fine-tuning. Inspired by ensemble learning, the decoder comprises a shared hierarchical layer followed by multiple parallel decoder branches, where each branch employs denoising queries either inherited from the shared layer or newly initialized to encourage prediction diversity. This design fully exploits pretrained weights without introducing additional parameters, and the resulting diverse predictions can be effectively ensembled to improve generalization. We further leverage a unified progressive fine-tuning framework with a plateau-aware learning rate schedule, which stabilizes optimization and achieves strong few-shot adaptation without complex data augmentations or extensive hyperparameter tuning. Extensive experiments on CD-FSOD, ODinW-13, and RF100-VL validate the effectiveness of our approach. Notably, on RF100-VL, which includes 100 datasets across diverse domains, our method achieves an average performance of 41.9 in the 10-shot setting, significantly outperforming the recent approach SAM3, which obtains 35.7. We further construct a mixed-domain test set from CD-FSOD to evaluate robustness to out-of-distribution (OOD) samples, showing that our proposed modules lead to clear improvement gains. These results highlight the effectiveness, generalization, and robustness of the proposed method. Code is available at: \url{https://github.com/Intellindust-AI-Lab/FT-FSOD}.
\end{abstract}

\vspace{-0.2cm}
\section{Introduction}
Few-shot object detection (FSOD) aims to detect novel object categories with only a few annotated examples. Beyond the challenge of limited data, a more practical difficulty arises when the downstream datasets with significant domain shifts from the natural images, such as industrial imagery and documentation contents.
To handle such domain shifts, recent progress in FSOD has been largely driven by the emergence of large-scale pretrained models~\cite{zhang2022glipv2,zhang2023detect,fu2024cross,liu2025dont,meng2025cdformer}, which provide transferable visual representations and enable rapid adaptation to new domains and categories.

Building on pretrained models, data generation and data augmentation searching methods~\cite{pan2025enhance,domainrag} enhance training data diversity and improve few-shot adaptation. While these strategies achieve impressive results, extensive computation is required, particularly on large and diverse benchmarks. Moreover, recent state-of-the-art (SOTA) results on large-scale FSOD benchmarks, including ODinW-13~\cite{zhang2022glipv2} and RF100-VL~\cite{robicheaux2025roboflow100vl}, are often achieved by fine-tuning large-parameterized foundation models such as SAM3~\cite{sam3_2025} and GroundingDINO 1.5 Pro~\cite{ren2024groundingpro}. These results highlight the strong transferability of large-scale pre-trained foundation models in few-shot settings. Nevertheless, effectively adapting pretrained detectors under significant domain shifts remains an active research problem.

\begin{figure}[t]
\centering
\includegraphics[width=1.0\linewidth]{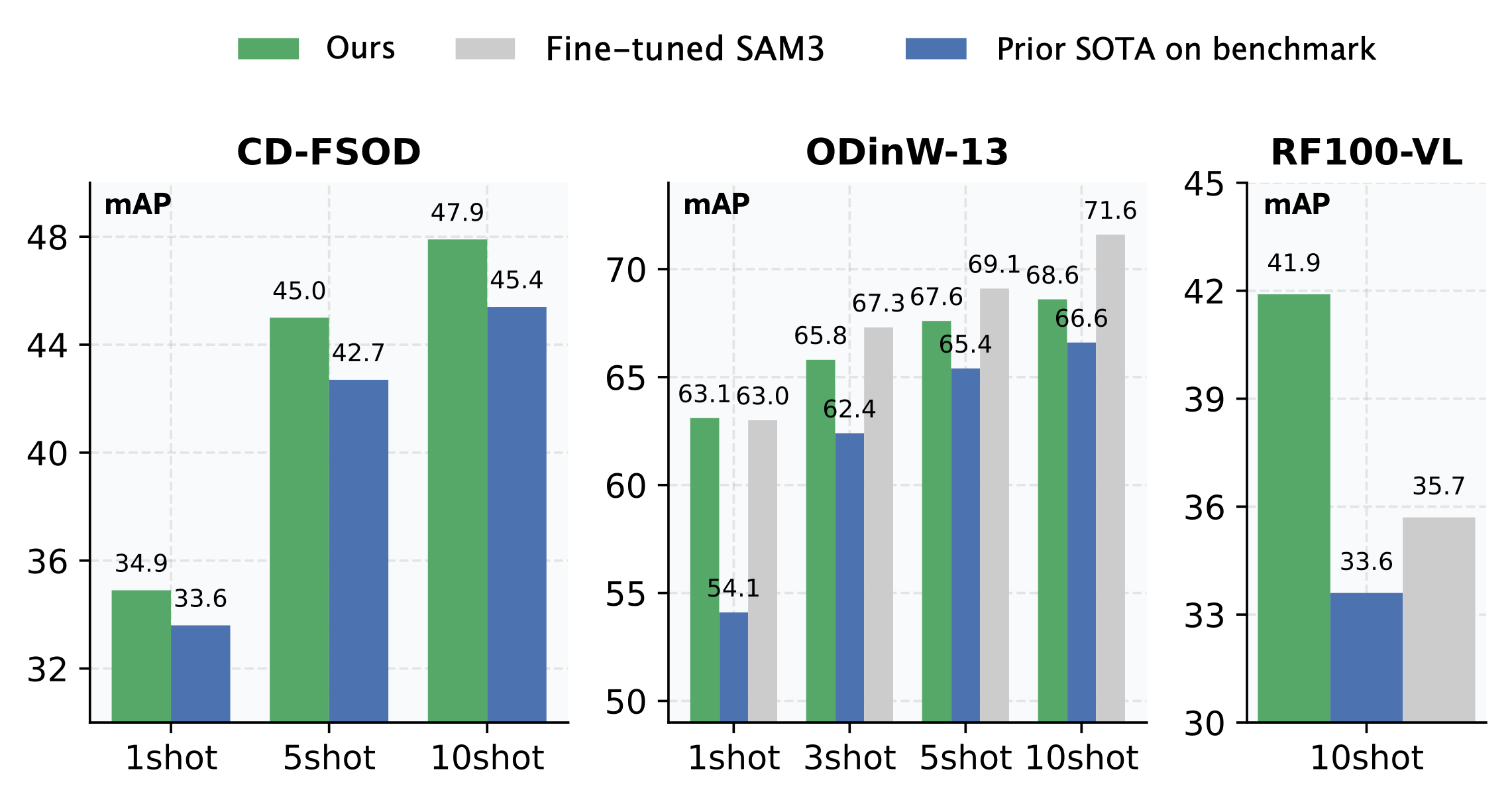}
\vspace{-0.5cm}
\caption{\textbf{Performance on large-scale cross-domain FSOD benchmarks.} All methods are adapted from pretrained models. Our approach, based on open-source MMGroundingDINO (MMGDINO)~\cite{zhao2024open}, surpasses prior SOTA (Domain-RAG~\cite{domainrag}, MQ-GLIP-L~\cite{mqdet2023} and MMGDINO-L~\cite{zhao2024open}) on CD-FSOD~\cite{fu2024cross}, ODinW-13~\cite{zhang2022glipv2} and RF100-VL~\cite{robicheaux2025roboflow100vl} benchmarks, and achieves comparable results to fine-tuned SAM3~\cite{sam3_2025}, notably outperforming it on the largest RF100-VL benchmark.}
\vspace{-0.2cm}
\label{fig:teaser}
\end{figure}

In this work, we take a closer look at the few-shot fine-tuning process of pretrained detectors from a simple and principled perspective. 
Rather than introducing additional data generation modules or complex architectural extensions, we focus on how to better utilize the intrinsic representation capacity of pretrained models during adaptation. Our goal is to explore whether consistent performance gains can be achieved through lightweight modifications and training strategies.
First, we introduce a hybrid ensemble decoder that integrates shared hierarchical layers with parallel decoder branches. 
During training, each branch uses denoising queries inherited from the shared layers or newly initialized, encouraging diverse predictions that are ensembled at inference.
Notably, this design fully reuses pretrained weights without introducing additional parameters. Second, we demonstrate that progressive fine-tuning yields substantial performance improvements. This unified, simple yet effective strategy requires no elaborate hyperparameter tuning for certain downstream datasets and remains robust even with basic data augmentations.

We evaluate our approach on three challenging cross-domain FSOD benchmarks: CD-FSOD~\cite{xiong2023cd}, ODinW-13~\cite{li2022grounded}, and RF100-VL~\cite{robicheaux2025roboflow100vl}. Results are illustrated in Fig.~\ref{fig:teaser}. On CD-FSOD, our method outperforms recent SOTA~\cite{pan2025enhance,domainrag} across 1/5/10-shot settings. It also surpasses the best-performed open-source model~\cite{mqdet2023} and remains competitive with other strong foundation models~\cite{sam3_2025,ren2024groundingpro} on ODinW-13 benchmark. Most notably, on the RF100-VL benchmark, which involves 100 heterogeneous downstream tasks, our method achieves an average score of 41.9, surpassing SAM3~\cite{sam3_2025}, which manages to obtain 35.7. These results highlight the adaptability and effectiveness of our approach, establishing a strong baseline in few-shot object detection.

Moreover, we evaluate the out-of-distribution (OOD) robustness of our approach. We utilize five datasets from CD-FSOD~\cite{fu2024cross} spanning diverse domains to construct a cross-domain mixed test set. 
In each evaluation round, one subset is treated as in-distribution with its corresponding fine-tuned model, while the remaining subsets serve as OOD samples. Results show that our hybrid ensemble decoder produces fewer overconfident predictions on OOD data, indicating improved robustness under distribution shifts.

In summary, our contributions are:
\begin{itemize}
\item We propose a hybrid ensemble decoder to improve feature diversity and enhance robustness against domain shifts without introducing extra parameters.
\item We introduce a progressive fine-tuning framework in few-shot scenarios, enabling consistent performance gains without complex hyperparameter tuning.
\item We achieve state-of-the-art performance on three representative benchmarks: CD-FSOD, ODinW-13, and RF100-VL, demonstrating the strong empirical effectiveness.
\item We conduct a comprehensive OOD analysis, validating the robustness of the proposed strategy under domain shifts and providing insights into its generalization behavior.
\end{itemize}

\section{Related works}\label{sec:Related_works}
\textbf{Object detection with transformers (DETR)~\cite{carion2020end}} formulates object detection as an end-to-end set prediction task, eliminating the need for NMS and marking a major paradigm shift from CNN-based detectors. Subsequent works have focused on accelerating convergence~\cite{zhu2020deformable,huang2025deim,jia2023detrs}, enhancing query design and matching strategies~\cite{liu2022dab,li2022dn,zhang2023dense}, and improving computational efficiency~\cite{zhao2024detrs,peng2024d}. Recently, DETR-based open-vocabulary object detection (OVOD) works such as GroundingDINO series~\cite{liu2024grounding,ren2024groundingpro,ren2024dino} demonstrate strong zero-shot detection capabilities. Notably, OVOD has been empirically shown to offer advantages for FSOD in recent studies~\cite{domainrag,madan2024revisiting,fu2025ntire,robicheaux2025roboflow100vl}. In this work, we build upon this observation and conduct a systematic investigation of few-shot fine-tuning strategies based on OVOD.

\noindent\textbf{Ensemble strategy in deep neural networks} is a long-standing approach to improve model accuracy and calibration of the predictions~\cite{dietterich2000ensemble,lakshminarayanan2017simple,dvornik2019diversity}, though traditional deep ensembles require training multiple independent models, leading to high computational costs. Recent studies propose more efficient solutions through implicit ensembling within a single network~\cite{havasi2021training,laurent2023packed}. In object detection, ensemble strategies are often realized via multi-head prediction or detector fusion~\cite{solovyev2021weighted,qutub2023bea} and Retentive R-CNN~\cite{fan2021generalized} also improve FSOD via detector head ensembles. Meanwhile, the ensemble technique can improve the OOD generalization ability and the robustness under distribution shifts~\cite{ovadia2019can,wenzel2020hyperparameter,gustafsson2020evaluating,yu2023robust}. Building on these insights, our Hybrid Ensemble Decoder (HED) implicitly ensembles parallel decoder layers with stochastic denoising query initialization, enhancing diversity and robustness without adding parameters or inference cost.

\begin{figure*}[t]
\centering
  \includegraphics[width=0.8\linewidth]{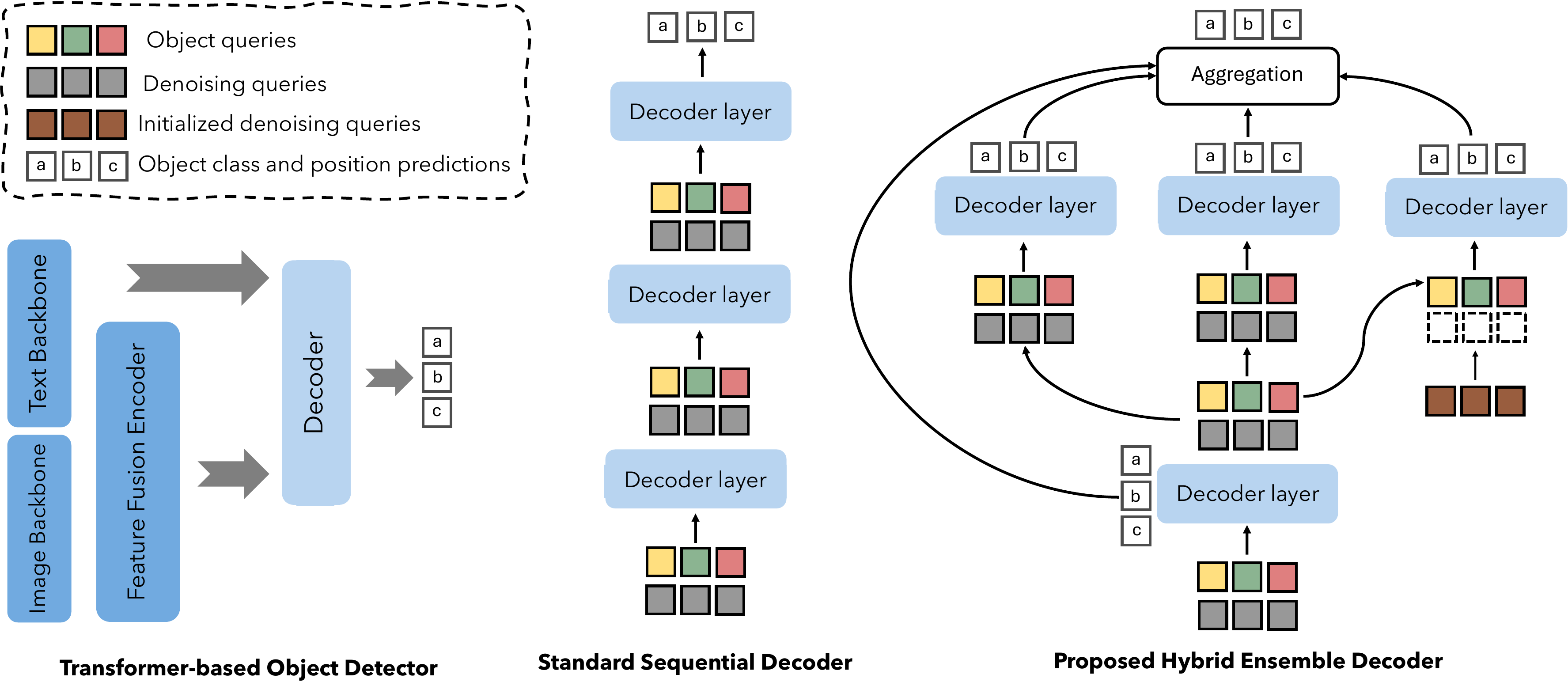}
  \vspace{-0.2cm}
  \caption{\textbf{Overview of the proposed hybrid ensemble decoder.} Our method extends the original decoder in a transformer-based object detector by parallelizing partial decoder layers on top of the model for the FSOD task. The final detection result is the aggregation of the results given by the object query outputs from all decoder layers. We also randomly replace the original denoising query with the initialized ones during training to further introduce diversity for the parallelized decoder layer input.}
  \label{fig:pipe} 
\vspace{-0.3cm}
\end{figure*}

\noindent\textbf{Few-shot object detection (FSOD)} aims to detect novel object categories with only a few labeled samples. Existing works mainly follow either meta-learning or transfer learning paradigms. The former learns transferable representations from {\tt base} classes for adaptation to {\tt novel} ones, using strategies such as feature re-weighting~\cite{kang2019few}, matching networks~\cite{fan2020few}, and prototype-based learning~\cite{wu2021universal}. In contrast, transfer learning methods pretrain on {\tt base} classes and fine-tune on $K$-shot {\tt novel} classes, yielding stronger performance~\cite{wang2020frustratingly}.
Recent advances leverage large-scale pretrained detectors and vision foundation models for FSOD~\cite{zhou2022detecting,zhang2023detect,fu2024cross,liu2025dont,meng2025cdformer,feng2026fsodvfm}, where ETS~\cite{pan2025enhance} and Domain-RAG~\cite{domainrag} further explore data augmentation during fine-tuning, yielding strong FSOD adaptability across domains.
In this work, we focus on the transfer learning paradigm and propose a unified progressive fine-tuning strategy that enhances FSOD generalization.
\section{Method}
In this section, we first review the Transformer-based object detection (DETR) paradigm~\cite{carion2020end} and its denoising variants~\cite{zhang2022dino,li2022dn,liu2024grounding} in Section~\ref{sec:detr}. These methods form the foundation of our hybrid ensemble decoder, introduced in Section~\ref{sec:ensemble}. The progressive training paradigm is introduced in Section~\ref{sec:prog_ft}, which enables more effective few-shot fine-tuning without adding extra architectural components.
\subsection{DETR preliminaries}
\label{sec:detr} 

\paragraph{Notations.} Let $\mathcal{D} = \{(x_i, \mathcal{Y}_i)\}_{i=1}^{N}$ denote a detection dataset with $N$ images,
where $x_i \in \mathbb{R}^{H \times W \times 3}$ represents the $i$-th input image, and
$\mathcal{Y}_i = \{(b_{ij}, c_{ij})\}_{j=1}^{M_i}$ denotes its corresponding set of ground truth objects with $M_i$ annotations.
Each annotation consists of a bounding box $b_{ij} \in \mathbb{R}^4$ and a category label $c_{ij} \in \{1, \dots, C\}$.
For clarity, we omit the image index $i$ in the following discussion.

\paragraph{DETR pipeline.}
Given an input image \(x\), a backbone network extracts feature maps 
\(f \in \mathbb{R}^{h \times w \times d}\), which are flattened and encoded by a transformer encoder into a sequence of visual tokens \(E\).
A set of \(N_q\) learnable queries \(Q^0 = \{q_i^0\}_{i=1}^{N_q}\) is then used by the \textit{standard sequential transformer decoder} (see the middle of Fig.~\ref{fig:pipe}) to interact with these tokens and predict objects.

At each of the \(L\) decoder layers, the queries are updated through self-attention (to reason among objects) and cross-attention (to attend to image features):
\begin{equation}
    Q^l = \text{DecoderLayer}^l(Q^{l-1}, E), \quad l \in \{1, \dots, L\}. 
    \label{eq:query}
\end{equation}
The final queries \(Q^L\) are fed into lightweight prediction heads to obtain class probabilities and bounding boxes. For the $i$-th query from $L$-layer $q_i^L$, the prediction is:
\[
\hat{p}_i = \text{ClsHead}(q_i^L), \qquad
\hat{b}_i = \text{BoxHead}(q_i^L).
\]

\paragraph{Objective function.}
DETR trains the decoder outputs to match the ground truth objects using a bipartite (Hungarian) matching strategy. 
Let the decoder produce a set of predictions 
\(\hat{\mathcal{Y}} = \{(\hat{b}_i, \hat{p}_i)\}_{i=1}^{N_q}\) for an input image,
where each query predicts a bounding box \(\hat{b}_i\) and a class score \(\hat{p}_i\).
The ground truth annotations of this image are 
\(\mathcal{Y} = \{(b_j, c_j)\}_{j=1}^{N}\).
Matching assigns each prediction to at most one ground truth object, forming pairs \((i, j) \in \pi\).

The training loss sums the classification and box regression terms over these matched pairs:
\begin{equation}
\label{eq:org_detection_loss}
\mathcal{L}_{\text{match}} =
\sum_{(i,j) \in \pi}
\big[
{\lambda_{\text{cls}}\,} \mathcal{L}_{\text{cls}}(\hat{p}_i, c_j)
+ \lambda_{\text{box}}\, \mathcal{L}_{\text{box}}(\hat{b}_i, b_j)
\big].
\end{equation}
Here, \(\mathcal{L}_{\text{cls}}\) classifies each query as one of the target categories, often using binary cross-entropy.
The box regression loss \(\mathcal{L}_{\text{box}}\) combines an \(\ell_1\) distance with the generalized IoU (GIoU)~\cite{rezatofighi2019generalized} term for stable localization, {and $\lambda_{\text{cls}}$ and $\lambda_{\text{box}}$ serve as hyperparameters.}
This learning objective and the bipartite matching enable DETR to be in a fully end-to-end manner without anchor design or heuristic assignments.

\paragraph{Training with denoising queries.}
While DETR-style detectors achieve fully end-to-end detection, they often converge slowly due to the sparse supervision and matching instability of the bipartite mechanism.
One solution to alleviate this issue is the \emph{denoising mechanism}~\cite{zhang2022dino,zhang2022dino,liu2024grounding}, which introduces additional queries that learn to recover groundtruth objects from intentionally corrupted versions and provides direct, stronger supervision to the decoder during training.  
The denoising branch is only applied during training and removed during inference, thus no extra computation is introduced during inference. 

Given the ground truth objects 
\(\mathcal{Y} = \{(b_j, c_j)\}_{j=1}^{N}\), a subset of boxes is randomly selected and slightly perturbed to form noisy targets:
\[
\tilde{b}_j = b_j + \delta_b, \qquad
\delta_b \sim \mathcal{U}(-\epsilon, \epsilon),
\]
while the category label \(\tilde{c}_j = c_j\) remains unchanged.  
Each noisy sample is converted into a denoising query by combining its box position and category embedding:
\[
q_j^{\text{dn}} = \text{Embed}(\tilde{b}_j, \tilde{c}_j).
\]
These queries are concatenated with the normal learnable queries before being processed by the decoder, allowing the model to learn how to correct the perturbations.

During training, the model predicts the clean boxes and labels \((b_j, c_j)\) from the noisy inputs.  
The denoising loss follows the same form as the standard DETR objective:
\begin{equation}
\label{eq:dn_loss}
\mathcal{L}_{\text{dn}} =
\sum_{(i,j)\in\pi_{\text{dn}}}
\Big[
\llf{\lambda_{\text{cls}}\,}\mathcal{L}_{\text{cls}}(\hat{p}_i, c_j)
+ \lambda_{\text{box}}\,\mathcal{L}_{\text{box}}(\hat{b}_i, b_j)
\Big],
\end{equation}
where \(\pi_{\text{dn}}\) represents the optimal matching between denoising predictions and their clean targets.
The overall training objective combines the standard detection and denoising losses introduced in Eq.~\ref{eq:org_detection_loss} and Eq.~\ref{eq:dn_loss}, respectively:
\begin{equation}
    \label{eq:total}
    \mathcal{L}_{\text{total}} =
\mathcal{L}_{\text{match}} + 
\lambda_{\text{dn}}\mathcal{L}_{\text{dn}}    
\end{equation}
with \(\lambda_{\text{dn}}\) controlling the weight of the denoising branch.

This auxiliary denoising task effectively strengthens the supervision signal by stabilizing the matching between object queries and ground truth, and involving additional supervision, thereby improving both convergence speed and final accuracy without altering inference-time complexity.

\subsection{Hybrid ensemble decoder}
\label{sec:ensemble} 
\paragraph{Hybrid ensemble decoder.} 
We propose a \emph{Hybrid Ensemble Decoder (HED)} that replaces the conventional fully sequential stack of decoder layers with a partially parallel structure, as shown on the right of Fig.~\ref{fig:pipe}. As defined in Eq.~\ref{eq:query}, a standard DETR decoder refines object queries sequentially across $L$ layers.  
In contrast, HED reorganizes these layers into two stages:  
\textit{(i)} the first $K$ layers follow the standard hierarchical refinement to produce stable and informative query features; and  
\textit{(ii)} the remaining $L\!-\!K$ layers are executed in parallel, each independently taking the output of the $K$-th layer, $Q^K$ as input. For the outputs of the parallel layers, we have:
\begin{equation}
    \label{eq:hed}
    Q^{K+m}=\text{DecoderLayer}^{K+m}(Q^K, E), m \in \{1, \dots, L-K\}.
\end{equation}
Each parallel branch thus develops a distinct variant of the query representation while reusing the pretrained decoder parameters.  
This design transforms the conventional refinement process into a lightweight ensemble of parallel decoder layer enabling greater diversity and better generalization without introducing additional parameters.

\paragraph{Ensemble-style aggregation.}
During inference, predictions from all decoder layers, both hierarchical and parallel, are aggregated to form the final outputs:
\begin{equation}
\label{eq:ensemble}
\hat{b} = \frac{1}{L} \sum\nolimits_{l=1}^{L} \hat{b}^l, \qquad
\hat{p} = \frac{1}{L} \sum\nolimits_{l=1}^{L} \hat{p}^l,
\end{equation}
where $\hat{b}^l$ and $\hat{p}^l$ denote the bounding box and classification outputs from the $l-th$ layer.  
This ensemble formulation transforms the traditional sequential refinement into an ensemble of query variants.  

Since each decoder layer in a pretrained DETR model has distinct weights, the parallel branches naturally behave as diverse sub-networks initialized from different points in parameter space, mirroring the principle of Deep Ensembles~\cite{lakshminarayanan2017simple}.  
However, when all parallel branches share identical input queries $Q^{K}$, their outputs may converge to similar solutions, limiting ensemble diversity.  
To mitigate this, we introduce input stochasticity using random denoising initialization.

\begin{table*}[t]
\centering
\scalebox{.76}{
\begin{tabular}{lllccccccc} 
\toprule
\multicolumn{1}{c}{} & \textbf{Method} & \multicolumn{1}{l}{\textbf{Backbone}} & \textbf{ArTaxOr} & \textbf{Clipart1k} & \textbf{DIOR} & \textbf{DeepFish} & \textbf{NEU-DET} & \textbf{UODD} & \textbf{Avg.} \\ 
\midrule
\textbf{0-shot} & MMGDINO-B~\cite{zhao2024open} & Swin-B~\cite{liu2021swin} & 7.3 & 48.8 & 5.3 & 42.0 & 1.2 & 20.0 & 20.8 \\
\midrule
\multirow{7}{*}{\rotatebox{90}{\textbf{1-shot}}} 
 & DE-ViT-FT$\dagger$~\cite{zhang2023detect} & DINOv2-L~\cite{oquab2024dinov2} & 10.5 & 13.0 & 14.7 & 19.3 & 0.6 & 2.4 & 10.1 \\
 & CD-ViTO~\cite{fu2024cross} & DINOv2-L~\cite{oquab2024dinov2} & 21.0 & 17.7 & 17.8 & 20.3 & 3.6 & 3.1 & 13.9 \\ 
 & VFMDETR~\cite{liu2025dont} & DINOv2-L~\cite{oquab2024dinov2} & 26.1 & 20.1 & 20.6 & 24.2 & 9.1 & 9.0 & 18.2 \\ 
 & CDFormer~\cite{meng2025cdformer} & DINOv2-L~\cite{oquab2024dinov2} & 36.0 & 54.0 & 16.3 & 34.5 & 7.4 & 12.7 & 26.8 \\ 
 & ETS~\cite{pan2025enhance} & Swin-B~\cite{liu2021swin} & 28.1 & \textbf{57.5} & 12.7 & \underline{40.7} & 11.7 & \underline{21.2} & 28.7 \\
 & Domain-RAG~\cite{domainrag} & Swin-B~\cite{liu2021swin} & \textbf{57.2} & \underline{56.1} & \underline{18.0} & 38.0 & \underline{12.1} & 20.2 & \underline{33.6} \\
 & \textbf{Ours} & Swin-B~\cite{liu2021swin} & \underline{49.1} & {55.6} & \textbf{24.6} & \textbf{42.7} & \textbf{15.5} & \textbf{22.1} & \textbf{34.9} \\ 
\midrule
\multirow{7}{*}{\rotatebox{90}{\textbf{5-shot}}} 
 & DE-ViT-FT$\dagger$~\cite{zhang2023detect} & DINOv2-L~\cite{oquab2024dinov2} & 38.0 & 38.1 & 23.4 & 21.2 & 7.8 & 5.0 & 22.3 \\
 & CD-ViTO~\cite{fu2024cross} & DINOv2-L~\cite{oquab2024dinov2} & 47.9 & 41.1 & 26.9 & 22.3 & 11.4 & 6.8 & 26.1 \\ 
 & VFMDETR~\cite{liu2025dont} & DINOv2-L~\cite{oquab2024dinov2} & 63.3 & 45.1 & 32.1 & 29.5 & 19.0 & 19.6 & 34.7 \\ 
 & CDFormer~\cite{meng2025cdformer} & DINOv2-L~\cite{oquab2024dinov2} & 65.0 & 58.9 & 28.1 & 31.7 & 15.0 & 23.8 & 37.1 \\ 
 & ETS~\cite{pan2025enhance} & Swin-B~\cite{liu2021swin} & 64.5 & \textbf{60.1} & 29.3 & 44.9 & 23.5 & \textbf{28.6} & 41.8 \\
 & Domain-RAG~\cite{domainrag} & Swin-B~\cite{liu2021swin} & \underline{70.0} & \underline{59.8} & \underline{31.5} & \underline{43.8} & \underline{24.2} & {26.8} & \underline{42.7} \\
  & \textbf{Ours} & Swin-B~\cite{liu2021swin} & \textbf{76.8} & {59.4} & \textbf{35.3} & \textbf{45.5} & \textbf{25.2} & \underline{27.5} & \textbf{45.0} \\ 
\midrule
\multirow{7}{*}{\rotatebox{90}{\textbf{10-shot}}} 
 & DE-ViT-FT$\dagger$~\cite{zhang2023detect} & DINOv2-L~\cite{oquab2024dinov2} & 49.2 & 40.8 & 25.6 & 21.3 & 8.8 & 5.4 & 25.2 \\
 & CD-ViTO~\cite{fu2024cross} & DINOv2-L~\cite{oquab2024dinov2} & 60.5 & 44.3 & 30.8 & 22.3 & 12.8 & 7.0 & 29.6 \\ 
 & VFMDETR~\cite{liu2025dont} & DINOv2-L~\cite{oquab2024dinov2} & 71.3 & 49.9 & 37.8 & 34.1 & 23.7 & 22.1 & 39.8 \\ 
 & CDFormer~\cite{meng2025cdformer} & DINOv2-L~\cite{oquab2024dinov2} & 68.7 & 59.0 & 32.5 & 35.5 & 18.1 & 26.4 & 40.0 \\ 
 & ETS~\cite{pan2025enhance} & Swin-B~\cite{liu2021swin} & 71.2 & \textbf{61.5} & 37.5 & \underline{44.1} & 26.1 & 29.8 & 45.0 \\
 & Domain-RAG~\cite{domainrag} & Swin-B~\cite{liu2021swin} & \underline{73.4} & \underline{61.1} & \underline{39.0} & {41.3} & \underline{26.3} & \underline{31.2} & \underline{45.4} \\
 & \textbf{Ours} & Swin-B~\cite{liu2021swin} & \textbf{79.2} & {59.6} & \textbf{41.5} & \textbf{46.3} & \textbf{28.4} & \textbf{32.3} & \textbf{47.9} \\
\bottomrule
\end{tabular}
}
\vspace{-0.1cm}
\caption{\textbf{Comparison of results (mAP$\uparrow$) of different models and strategies on the CD-FSOD~\cite{xiong2023cd} benchmark under the 1/5/10-shot setting.} Best and second-best results are highlighted in bold and underlined, respectively. $\dagger$ marks the results reproduced in CD-ViTO~\cite{fu2024cross}.}
\label{tab:maintable_cdfsod}
\vspace{-0.3cm}
\end{table*}

\paragraph{Random denoising query initialization.}
In DETR variants with denoising mechanisms, each decoder layer receives two types of queries: object queries and denoising queries, both propagated sequentially from the previous layer.  
In our hybrid design, the first $K$ layers follow this standard refinement.  
For the parallel layers, we inject controlled randomness into the denoising branch to increase diversity.  
Given the denoising queries $Q_{\text{dn}}^{K}$ from the $K$-th layer, we randomly re-initialize them with probability $\tau \in [0,1]$ before feeding them into each parallel branch:
\begin{equation}
\label{eq:reinit_dnq}
Q_{\text{dn}}^{K+m} =
\begin{cases}
\text{RandInit}, & \text{w. prob. } \tau, \\
Q_{\text{dn}}^{K}, & \text{otherwise},
\end{cases}
\quad m = 1, \dots, L-K.
\end{equation}
When $\tau = 0$, all parallel branches receive the same inputs, reverting to the standard formulation.  
When $\tau > 0$, partially initialized denoising queries involve input diversity, while object queries remain clean and unchanged to preserve semantic stability.

The training for the proposed architecture still follows the overall loss in Eq.~\ref{eq:total}. For each parallel branch, denoising targets in Eq.~\ref{eq:dn_loss} are redefined according to its random initialization, and losses are averaged across branches to form a unified supervision signal.  
Since denoising is used only during training, inference remains deterministic and stable.

\paragraph{Discussion.}
The proposed Hybrid Ensemble Decoder (HED) is especially effective in few-shot scenarios, where limited data often leads to overfitting.  
By reusing pretrained decoder layers without adding parameters, HED retains the inductive bias of the pretrained model while remaining efficient.  
The parallel branches introduce prediction diversity and improve generalization, similar to the effect of promising ensemble strategies ~\cite{lakshminarayanan2017simple,laurent2023packed}.

We further observe that a purely parallel configuration (\ie, $K=0$) performs worse, indicating the necessity of at least one hierarchical stage to produce stable, object-aware query features.  
Empirically, combining outputs from both hierarchical and parallel stages achieves the best performance (see Sec.~\ref {sec:ablation_study}), as they contribute complementary strengths: hierarchical layers maintain semantic consistency, while parallel branches \and the randomly initialized denoising queries enhance robustness through diversity, see Sec.~\ref {sec:robust_exp}.

In summary, HED reframes the DETR decoder as a structured ensemble that balances stability and diversity, improving robustness and generalization without increasing parameters or inference cost.

\begin{table*}[t]
\centering
\scalebox{.74}{
\begin{tabular}{llcccccc} 
\toprule
\textbf{Model} & \textbf{Backbone} & \textbf{Ckpt avail.} & \textbf{0-shot} & \textbf{1-shot} & \textbf{3-shot} & \textbf{5-shot} & \textbf{10-shot} \\ 
\midrule
GLIPv2-H~\cite{zhang2022glipv2} & Swin-H~\cite{liu2021swin} & \XSolidBrush & 55.5 & $61.7 \pm 0.5$ & $64.1 \pm 0.8$ & $64.4 \pm 0.6$ & $65.9 \pm 0.3$ \\
GLEE-Pro~\cite{wu2024glee} & EVA02-L~\cite{fang2024eva} & \Checkmark & 53.4 & $59.4 \pm 1.5$ & $61.7 \pm 0.5$ & $64.3 \pm 1.3$ & $65.6 \pm 0.4$ \\
MQ-GLIP-L~\cite{mqdet2023} & Swin-L~\cite{liu2021swin} & \Checkmark & 54.1 & 62.4 & 64.2 & 65.4 & 66.6 \\
Grounding DINO 1.5 Pro~\cite{ren2024groundingpro} & EVA02-L~\cite{fang2024eva} & \XSolidBrush & 58.7 & $62.4 \pm 1.1$ & $\underline{66.3} \pm 1.0$ & $66.9 \pm 0.2$ & $67.9 \pm 0.3$ \\
\color{gray} SAM3~\cite{sam3_2025} & \color{gray} PE-L+~\cite{bolya2025perception} & \color{gray} \Checkmark & \color{gray} 59.9 & \color{gray} $63.0 \pm 2.4$ & \color{gray} $\textbf{67.3} \pm 0.7$ & \color{gray} $\textbf{69.1} \pm 1.1$ & \color{gray} {$\textbf{71.6} \pm 0.2$} \\
\textbf{Ours} (MMGDINO-L~\cite{zhao2024open}) & Swin-L~\cite{liu2021swin} & \Checkmark & 55.3 & $ \textbf{63.1} \pm 0.6 $ & $ 65.8 \pm 0.7 $  & $ \underline{67.6} \pm 0.3 $ & $ \underline{68.6} \pm 0.6 $ \\
\bottomrule
\end{tabular}
}

\footnotesize \color{gray}{Gray} indicates that the result was not obtained via conventional training-validation model selection.
\vspace{-2mm}
\caption{\textbf{Comparison of results (mAP$\uparrow$) of different models and strategies on ODinW-13~\cite{li2022grounded} benchmark} under 1/3/5/10-shot setting. 0-shot performance is for reference. The results are averaged by {three} official seeds of runs. Best and second-best results are highlighted in bold and underlined, respectively.}
\vspace{-0.3cm}
\label{tab:maintable_odinw}
\end{table*}

\begin{table*}[t]
\centering
\scalebox{.75}{
\begin{tabular}{llcccccccc} 
\toprule
\textbf{\textbf{Model}} & \textbf{Backbone} & \textbf{Aerial} & \textbf{Document} & \textbf{Flora-Fauna} & \textbf{Industrial} & \textbf{Medical} & \textbf{Sports} & \textbf{Other} & \textbf{Avg.} \\ 
\midrule
Detic~\cite{zhang2023detect} & Swin-L~\cite{liu2021swin} & 12.2 /~19.5 & 4.5 /~19.6 & 17.9 /~28.4 & 6.0 /~25.9 & 0.8 /~8.5 & 7.6 /~26.6 & 11.2 /~25.7 & 9.5 /~22.8 \\
\color{gray} SAM3~\cite{sam3_2025} & \color{gray} PE-L+~\cite{bolya2025perception} & \color{gray} 20.7 /~\underline{33.9} & \color{gray} 11.9 /~\underline{34.0} & \color{gray} 23.4 /~38.8 & \color{gray} 8.2 /~\underline{39.7} & \color{gray} 2.0 /~\underline{24.9} & \color{gray} 16.4 /~\textbf{40.3} & \color{gray} 15.7 /~\underline{34.7} & \color{gray} 14.3 /~\underline{35.7} \\
MMGDINO-L~\cite{zhao2024open} & Swin-L~\cite{liu2021swin} & 21.5 /~32.4 & 9.2 /~30.6 & 27.9 /~\underline{41.3} & 10.3 /~37.8 & 2.1 /~18.3 & 13.3 /~33.2 & 17.5 /~32.0 & 15.7 /~33.6 \\
\textbf{Ours} (MMGDINO-L) & Swin-L~\cite{liu2021swin} & 21.5 /~\textbf{41.5} & 9.2 /~\textbf{38.9} & 27.9 /~\textbf{45.9} & 10.3 /~\textbf{46.7} & 2.1 /~\textbf{25.8} & 13.3 /~\underline{36.2} & 17.5 /~\textbf{47.2} & 15.7 /~\textbf{41.9} \\
\bottomrule
\end{tabular}
}

\footnotesize \color{gray}{Gray} indicates that the result was not obtained via conventional training-validation model selection.
\vspace{-2mm}
\caption{\textbf{Comparison of results (mAP$\uparrow$) of different models and strategies on RF100-VL~\cite{robicheaux2025roboflow100vl} benchmark}. The results are illustrated as {0/10-shot} setting. 0-shot performance is for completeness. Best and second-best results are highlighted in bold and underlined, respectively.}
\vspace{-0.35cm}
\label{tab:maintable_rf100}
\end{table*}

\subsection{Unified progressive fine-tuning framework}
\label{sec:prog_ft}
\paragraph{Dataset-agnostic augmentation and adaptive learning rate scheduling.} Few-shot datasets span diverse domains with distinct visual characteristics, making hyperparameter and augmentation tuning particularly sensitive. Instead of performing dataset-specific data augmentation searching~\cite{pan2025enhance}, which is often computationally expensive, we adopt a unified yet robust training pipeline that generalizes well across domains. While Domain-RAG~\cite{domainrag} employs FLUX~\cite{fluxfill2024} for generative data augmentation, such approaches are resource-intensive, and edited images cannot be guaranteed to be free of subtle artifacts that are visually imperceptible in the spatial domain but can still influence training, as discussed in~\cite{wang2020high}.
We argue that all datasets have their conductive bias, thus our augmentation pipeline relies on simple, stable operations, such as random flipping, color jittering, and random mixup~\cite{zhang2017mixup}, that consistently improve diversity without requiring per-dataset adjustment.

To further avoid dataset-specific hyperparameter tuning, we employ a plateau learning rate scheduler~\cite{paszke2019pytorch} that automatically adapts to the convergence behavior of each dataset by adjusting the learning rate according to the validation performance.
This design enables consistent, resource-efficient adaptation across heterogeneous few-shot benchmarks. More training details are provided in the Appendix.

\vspace{-0.2cm}
\paragraph{Progressive few-shot fine-tuning.} 
Adapting large pretrained detectors to few-shot data is challenging, as direct full fine-tuning often leads to overfitting.
We observe that a progressive fine-tuning strategy effectively mitigates these issues, similar ideas are also explored in previous works such as DeFRCN~\cite{qiao2021defrcn} and LP-FT~\cite{kumar2022finetuning}. Specifically, we adopt a two-stage training paradigm: In the first stage, we freeze the encoder, and in the second stage, we unfreeze all parameters for full fine-tuning. The transition between stages is automatically triggered when the learning rate decays under the plateau scheduler introduced in the previous section, ensuring that the switch occurs only after the model has sufficiently stabilized.

\section{Experiments}
\subsection{Benchmarks and base models}
We conduct experiments on three representative few-shot detection benchmarks: \textit{CD-FSOD}~\cite{fu2024cross}, \textit{ODinW-13}~\cite{li2022grounded}, and \textit{RF100-VL}~\cite{robicheaux2025roboflow100vl}, which together span diverse domains and data scales. \textit{CD-FSOD} consists of 
6 datasets with 1/5/10-shot settings and provides only few-shot training sets with full-data test sets. \textit{ODinW-13} covers 13 datasets under 1/3/5/10-shot settings, each including few-shot training and validation splits. Best models are selected based on validation sets before evaluating on full-data test sets. \textit{RF100-VL} further introduces cross-domain few-shot object detection across 100 datasets, offering 10-shot training/validation splits and full-data test sets. The best checkpoints are also selected based on the few-shot validation sets.

Following recent studies~\cite{domainrag,madan2024revisiting,fu2025ntire,robicheaux2025roboflow100vl}, we adopt open-source open-vocabulary detectors as our base models. Specifically, for CD-FSOD~\cite{fu2024cross}, we use MMGroundingDINO-B (MMGDINO-B)~\cite{zhao2024open}, consistent with the recent SOTA Domain-RAG~\cite{domainrag}. For RF100-VL~\cite{robicheaux2025roboflow100vl}, we use larger size MMGDINO-L~\cite{zhao2024open}, which follows the standard baseline provided in this benchmark~\cite{robicheaux2025roboflow100vl}. For ODinW-13~\cite{li2022grounded}, we also use MMGDINO-L, which shows reasonable zero-shot performance and represents the limit of our computational resources. While more competitive approaches, such as SAM3~\cite{sam3_2025} and GroundinDINO 1.5 Pro~\cite{ren2024groundingpro}, employ larger models trained on significantly larger datasets, our goal is to push the performance of publicly available models within a reasonable computational budget. 
Similarly, for ODinW-13~\cite{li2022grounded} we use MMGDINO-L as a comparable alternative with similar zero-shot capability. We note that SAM3 implementations do not use dedicated few-shot validation sets for model selection during training. As a result, comparisons involving SAM3 are reported separately.
Overall, our choices ensure a fair and reproducible evaluation for few-shot adaptation. Due to space limitations, the results for sub-datasets in ODinW-13 and RF100-VL and the implementation details for different base models will be provided in the Appendix.

\subsection{Comparison to competitive approaches}
\paragraph{Results of CD-FSOD~\cite{xiong2023cd}.}
We evaluate model performance across different datasets and few-shot settings, and the complete results are presented in Tab.~\ref{tab:maintable_cdfsod}. Our method achieves state-of-the-art results across the 1-, 5-, and 10-shot settings using the same pretrained model, without introducing any additional data or auxiliary models. We observe that Domain-RAG~\cite{domainrag} and ETS~\cite{pan2025enhance} perform well on datasets such as Clipart1k, which involve detecting common objects (e.g., people, cars) in cartoon-like images, where generative data augmentation or augmentation searching can provide additional domain coverage. However, their improvement is marginal in more specialized domains such as NEU-DET, which focuses on industrial defect detection.
In contrast, our approach maintains consistently strong performance across both common and rare domains, demonstrating its capability without relying on domain-specific generation.

\paragraph{Results of ODinW-13~\cite{li2022grounded}.}
We report results in Tab.~\ref{tab:maintable_odinw}. This benchmark is commonly used to evaluate the generalization capability of open-vocabulary detectors. We compare our method against recent large-scale pretrained detectors, including the concurrent SAM3~\cite{sam3_2025} and GDINO 1.5 Pro~\cite{ren2024groundingpro}, which are exceptionally strong competitors with more powerful backbones and extensive pretraining data. Despite the 0-shot performance of our base model being worse, our few-shot results show substantial gains. 
Using the pretrained 
MMGDINO-L~\cite{zhao2024open}, our method outperforms or is comparable to the results from both open-source and closed-source models, highlighting the effectiveness of the approach.

\paragraph{Results on RF100-VL~\cite{robicheaux2025roboflow100vl}.}
Tab.~\ref{tab:maintable_rf100} presents the performance comparison between our method and state-of-the-art detectors on RF100-VL, the largest and most diverse few-shot object detection benchmark, containing 100 datasets spanning a wide range of domains. The zero-shot results of our base model are generally modest across different datasets, but applying the proposed few-shot fine-tuning strategy leads to almost all subsets surpassing the previous SOTA results, demonstrating the effectiveness of our approach in adapting pretrained models to new domains. Importantly, our training strategy is fully automated and avoids the need for manual hyperparameter tuning, a critical advantage when working with 100 heterogeneous datasets, where per-dataset tuning would be infeasible. These results highlight both the robustness and generalization capability of our approach.

\begin{table}[t]
\centering
\scalebox{.77}{
\begin{tabular}{llcccc} 
\toprule
\multicolumn{2}{l}{\textbf{Baseline (Naive aug \& Plateau scheduler)}} & \Checkmark & \Checkmark & \Checkmark & \Checkmark \\
\multicolumn{2}{l}{\textbf{Progressive Fine-tuning}} &  &  \Checkmark &   & \Checkmark \\
\multicolumn{2}{l}{\textbf{Hybrid Ensemble Decoder}} &  &  & \Checkmark  & \Checkmark \\ 
\midrule
\multirow{3}{*}{\begin{tabular}[c]{@{}l@{}}\textbf{\textbf{CD-FSOD}} \\(0-shot: 20.8)\end{tabular}} & \textbf{1-shot} & 30.8 & 33.3 & 33.1 & \textbf{34.9} \\
 & \textbf{5-shot} & 43.1 & 44.6 & 43.4 & \textbf{45.0} \\
 & \textbf{10-shot} & 47.1 & 47.1 & 47.2 & \textbf{47.9} \\ 
\midrule
\multirow{3}{*}{\begin{tabular}[c]{@{}l@{}}\textbf{\textbf{ODinW-13}} \\(0-shot: 55.3)\end{tabular}} & \textbf{1-shot} & 60.7 & 61.8 & 62.0 & \textbf{63.1} \\
 & \textbf{5-shot} & 65.4 & 65.8 & 65.3 & \textbf{67.6} \\
 & \textbf{10-shot} & 67.2 & 68.1 & 67.0 & \textbf{68.6} \\
\bottomrule
\end{tabular}

}
\vspace{-0.1cm}
\caption{\textbf{Ablation study on the proposed approach on CD-FSOD~\cite{xiong2023cd} and ODinW-13~\cite{li2022grounded} under 1/5/10-shot settings.} 
The results are averaged by three official seeds of runs on ODinW-13. Best results are highlighted in bold.}
\vspace{-0.3cm}
\label{tab:ablation_training}
\end{table}

\begin{figure*}[t]
\centering
  \includegraphics[width=0.9\linewidth]{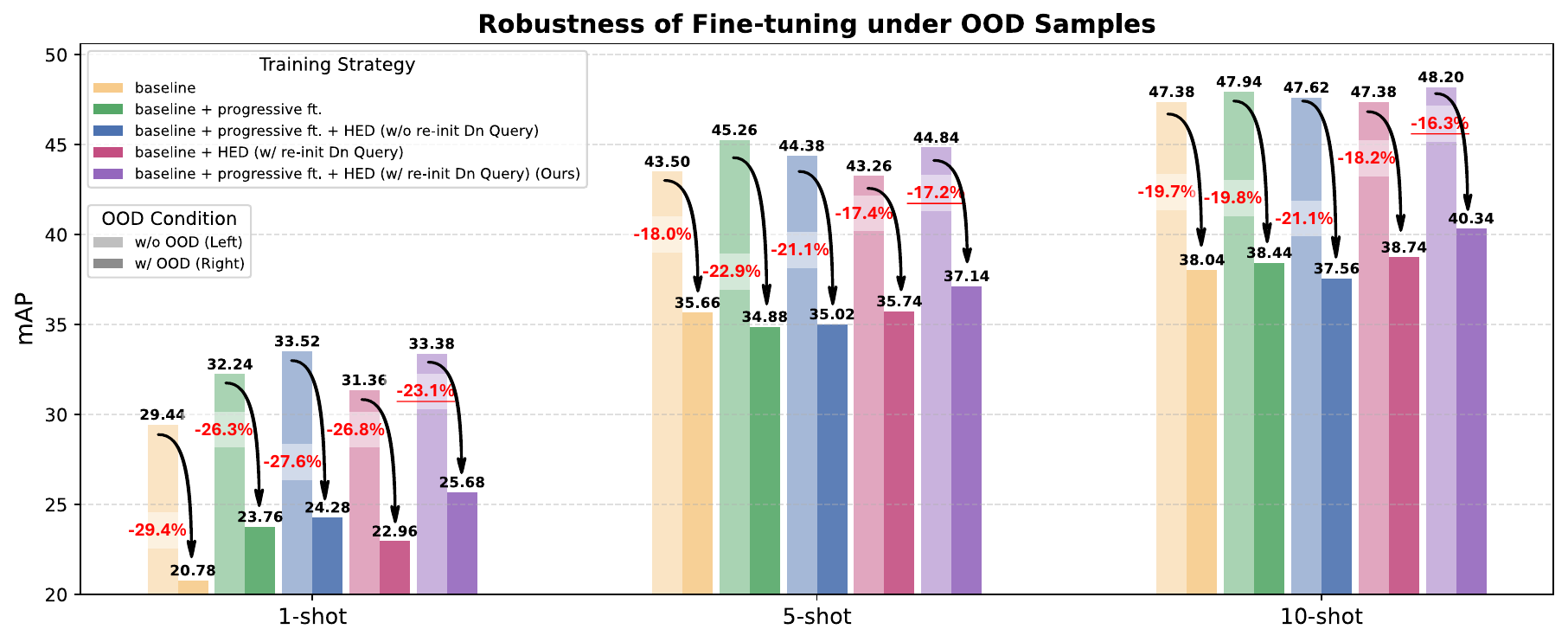}
  \vspace{-0.2cm}
  \caption{\textbf{Illustration on performance reduction of different fine-tuning strategies when the test set contains OOD samples.} The performance reductions are highlighted in bold red, and the most robust results are underlined.}
  \label{fig:robustness_results} 
\vspace{-0.2cm}
\end{figure*}

\subsection{Ablation study}
\label{sec:ablation_study}
We conduct ablation experiments to study how proposed strategies work and the impact on the involved hyperparameters. These experiments are based on CD-FSOD~\cite{fu2024cross} and ODinW-13~\cite{li2022grounded} datasets. 
To be consistent with Tab.~\ref{tab:maintable_cdfsod} and Tab.~\ref{tab:maintable_odinw}, we still adopt 
MMGDINO-B~\cite{zhao2024open} for CD-FSOD and MMGDINO-L for ODinW-13 for all ablation experiments. Other ablations on learning rate scheduler and decoder initialization are in the Appendix.

\paragraph{Ablation study and effectiveness of the proposed components.}
We conduct comprehensive ablation studies on the proposed modules across CD-FSOD~\cite{fu2024cross} and ODinW-13~\cite{li2022grounded} under 1/5/10-shot settings, as summarized in Tab.~\ref{tab:ablation_training}. 
The results reveal several key observations.
\textit{(i)} Combining simple data augmentation with a plateau learning rate scheduler already provides a strong baseline, effectively reducing the need for delicate hyperparameter tuning.
\textit{(ii)} Both the proposed progressive fine-tuning and the hybrid ensemble decoder (HED) yield consistent improvements over this baseline, and are already comparable to or surpass the SOTA~\cite{domainrag}, confirming their individual effectiveness. Furthermore, these two components exhibit strong complementarity. The two-stage fine-tuning enhances optimization stability and mitigates overfitting, while HED improves generalization by implicitly ensembling sub-networks in the decoder.

Notably, the performance gain differs across benchmarks. On CD-FSOD, where the zero-shot baseline is relatively weak, both components bring substantial improvements. In contrast, ODinW-13 already demonstrates strong zero-shot generalization, leaving less room for fine-tuning gains. Nevertheless, combining the progressive fine-tuning with HED leads to the most stable and consistent performance improvements across all shot settings. Overall, the results confirm that these designs jointly contribute to more effective few-shot adaptation and achieve competitive SOTA performance.

\paragraph{Ablation on the hybrid ensemble decoder (HED).}
We conduct an ablation study to analyze how different design choices affect the performance of the proposed HED, as summarized in Tab.~\ref{tab:ablation_parallelization}. Due to limited computational resources, experiments on ODinW-13~\cite{li2022grounded} are performed with one official seed. Specifically, we vary the ratio of stacked to parallel decoder layers and the re-initialization rate $\tau$ of denoising queries. All models are trained under the proposed progressive fine-tuning framework.

We observe that introducing decoder-layer parallelism improves performance compared to the baseline, except for the fully parallel (\textit{6-parallel}) setup. This suggests that the hybrid ensemble of stacked and parallel decoders is beneficial and that the initial refinement of object queries is essential. Among all configurations, the \textit{1-stacked + 5-parallel} structure with $\tau=0.5$ for denoising queries achieves more balanced performance across datasets. Hence, we adopt this setup for all large benchmark experiments shown in Tab.~\ref{tab:maintable_cdfsod}~\ref{tab:maintable_odinw}~\ref{tab:maintable_rf100}. We also argue that varying the re-initialization rate of denoising queries leads to relatively stable in-distribution accuracy. Yet, as will be discussed in Sec.~\ref{sec:robust_exp}, setting $\tau$ to zero eliminates input diversity, resulting in overconfidence on OOD samples. Overall, both decoder parallelization and denoising query design play crucial roles in enhancing the effectiveness of HED for few-shot object detection.

\begin{table}[t]
\centering
\scalebox{.75}{
\begin{tabular}{lrcccc} 
\toprule
\multirow{2}{*}{\begin{tabular}[c]{@{}l@{}}\textbf{Decoder}\\\textbf{\textbf{\textbf{\textbf{\textbf{\textbf{\textbf{\textbf{Structure}}}}}}}}\end{tabular}} & \multicolumn{1}{l}{\multirow{2}{*}{\begin{tabular}[c]{@{}l@{}}\textbf{DnQ re-init}\\\textbf{rate $\tau$ (Eq.~\ref{eq:reinit_dnq})}\end{tabular}}} & \multicolumn{2}{c}{\textbf{CD-FSOD}} & \multicolumn{2}{c}{\textbf{ODinW-13}} \\ 
\cmidrule(l){3-6}
 & \multicolumn{1}{l}{} & \textbf{1-shot} & \textbf{5-shot} & \textbf{1-shot} & \textbf{5-shot} \\ 
\midrule
\multicolumn{2}{l}{6-stacked (Baseline+prog. ft.)~~~~~~~ --} & 33.3 & 44.6 & 61.8 & 66.1 \\ 
\midrule
6-parallel & 0 & 27.4 & 40.0 & 58.4 & 63.6 \\
4-stacked+2-parallel & 0 & 33.9 & 44.8 & 60.3 & 66.6 \\
3-stacked+3-parallel & 0 & 32.9 & 44.3 & 60.0 & \textbf{67.3} \\
2-stacked+4-parallel & 0 & 34.2 & 44.3 & 61.9 & 66.4 \\
1-stacked+5-parallel & 0 & \textbf{\textbf{35.1}} & 44.6 & 62.8 & 66.5 \\ 
\midrule
1-stacked+5-parallel & 0.25 & 34.7 & 44.7 & 62.5 & 66.4 \\
1-stacked+5-parallel & 0.50 & 34.9 & \textbf{45.0} & \textbf{63.4} & 67.1 \\
1-stacked+5-parallel & 0.75 & 34.8 & 44.8 & 63.1 & 66.2 \\
1-stacked+5-parallel & 1.00 & 34.3 & 44.4 & 63.3 & 66.7 \\
\bottomrule
\end{tabular}
}
\vspace{-0.1cm}
\caption{\textbf{Ablation study on proposed hybrid ensemble decoder.} Experiments are conducted on CD-FSOD~\cite{xiong2023cd} and ODinW-13~\cite{li2022grounded} (only on seed3 subset) 1/5-shot settings . Best results are in bold.}
\label{tab:ablation_parallelization}
\vspace{-0.3cm}
\end{table}

\subsection{Towards more reliable CD-FSOD fine-tuning}
\label{sec:robust_exp}
In real-world semi-automated labeling, the fine-tuned models often encounter out-of-distribution (OOD) samples, \ie, images without target objects and from different domains. This challenge is commonly studied under the settings of open-set object detection~\cite{dhamija2020overlooked} or evaluated through false-positive robustness benchmarks~\cite{liu2024coco}. A robust and reliable detector should perform well on target images while producing low-confidence predictions on OOD samples, as overconfident errors can reduce final mAP.
\paragraph{Experiment settings.}To evaluate the robustness of our fine-tuning strategy under this OOD condition, similar to COCO-FP~\cite{liu2024coco}, we construct a cross-domain mixed test set, dubbed \textit{CD-Mixed set}, by combining images from the test set of five distinct datasets in CD-FSOD benchmark~\cite{fu2024cross}, namely, ArTaxOr, Clipart1k, DIOR, NEU-DET, and UODD, respectively contain close-ups of the insects, clipart drawings, aerial scenes, industrial defects, and underwater organisms. FISH dataset is not included since its domain overlaps with UODD. The test sets in CD-Mixed set have completely disjoint object categories and domain characteristics.

We perform five evaluation rounds on the CD-Mixed set, where in each round, the ground truth is taken from one of the constituent datasets and the model is fine-tuned on the corresponding training set, while the other datasets serve as OOD samples. The final result is the average mAP according to the five rounds of evaluation. This setup allows us to examine whether different fine-tuning strategies generate overconfident false-positive predictions when faced with a large number of target-irrelevant images and various domain transformations, \ie, the robustness to the OOD samples.

\paragraph{Results on CD-Mixed set.}The results are illustrated in Fig.~\ref{fig:robustness_results} {and the full numerical results are shown in the Appendix}. We observe that the performance of all fine-tuning strategies degrades when OOD samples are introduced, indicating that the fine-tuned detectors tend to be overconfident. Moreover, as the number of training shots decreases, models become more vulnerable to OOD contamination. 

We further find that both the progressive fine-tuning and naive decoder parallelism can moderately improve detection accuracy under both in-distribution and OOD settings. However, regarding the magnitude of mAP degradation, neither approach effectively prevents the model from being overconfident on OOD samples. By contrast, when input diversity is introduced, \ie, random initialization for denoising queries, the performance reduction becomes consistently smaller, regardless of whether progressive fine-tuning is applied. This suggests that combining input diversity and decoder layer ensembles helps reduce overconfidence and enhances generalization.
Overall, combining progressive fine-tuning with the hybrid ensemble decoder (HED) yields the most reliable and robust detector, maintaining high accuracy while effectively mitigating OOD overconfidence.

\section{Conclusions}
In this work, we address cross-domain few-shot object detection with a hybrid ensemble decoder (HED) and a unified progressive fine-tuning strategy. HED integrates shared hierarchical layers with parallel decoders using randomly inherited and re-initialized denoising queries, enhancing generalization through implicit ensembling without extra parameters. The unified progressive fine-tuning scheme further stabilizes optimization and improves few-shot adaptation without extensive hyperparameter searching or complicated data augmentation. 
Extensive experiments on CD-FSOD, ODinW-13, and RF100-VL demonstrate strong performance, while evaluations on a mixed-domain test set show better robustness to out-of-distribution samples. Overall, our approach offers a simple yet effective framework for stable and generalizable few-shot detection.

\section*{Acknowledgments}
\begin{flushleft}
This work is supported by the NSF of China under Grant~62502492.
\end{flushleft}
{
    \small
    \bibliographystyle{ieeenat_fullname}
    \bibliography{main}
}

\clearpage

\setcounter{page}{1}
\maketitlesupplementary

\appendix
\renewcommand{\thesection}{\Alph{section}}

\startcontents
\printcontents{}{1}{\section*{Contents}}

\section{Full training details}

\begin{table*}[h]
\centering
\begin{tabular}{ll} 
\toprule
\textbf{Component} & \textbf{Hyperparameters} \\ 
\midrule
Init. learning-rate for detection transformers & 1e-4 \\
Init. learning-rate for image and language backbones & 2e-5 \\
Minimum learning rate & 1e-6 \\
Weight decay & 0.05 \\
Batch size & 4 \\
Number of epochs & 100 \\
Plateau scheduler - Patience w/o progressive ft. & 5 \\
Plateau scheduler - Patience in the~1st stage w/ progressive ft. & 3 \\
Plateau scheduler - Patience in the 2nd stage w/ progressive ft. & 8 \\
Plateau scheduler - factor & 0.5 \\
Re-initialization ratio for denoising queries & 0.5 \\
Hybrid ensemble decoder structure & 1-stacked layer + 5-parallel layers \\
Random flip probability & 0.5 \\
YOLOXHSVRandomAug probability & 0.5 \\
CatchedMixup probability & 0.3 \\
\bottomrule
\end{tabular}
\caption{\textbf{Hyperparameter list applied for training on all benchmark experiments.}}
\label{tab:supp_param}
\end{table*}

\paragraph{Hyperparameters.} We list the hyperparameters of our training in Table~\ref{tab:supp_param}. Specifically, patience without progressive fine-tuning in the plateau scheduler indicates that the patience used in the \textit{Baseline (Naive aug \& Plateau scheduler)} in the experiments in Table~\ref{tab:ablation_training}, as well as the \textit{baseline} and the \textit{baseline + HED (w/ re-init Dn Query)} in Figure~\ref{fig:robustness_results} and Table~\ref{tab:supp_cdmix}. For the progressive fine-tuning, we set different patience for the first and second stages. The smaller patience in the first stage can prevent fast overfitting caused by the bigger learning rate at the beginning of the training, and the bigger patience in the second stage with a smaller learning rate can help the model to retain the good performance after the first plateau. When the model is trained without the progressive fine-tuning pipeline, the patience is set to the average patience for the whole training as a reasonable configuration. {Note that we train 100 epochs for the sub-datasets, which is the same configuration as RF-DETR~\cite{robinson2026rfdetr} used on RF100-VL full-shot benchmark~\cite{robicheaux2025roboflow100vl}, in order to make sure that the models are correctly and fully converged. This does not indicate that training all datasets requires this configuration, and one can set early stopping to save time.}
We apply the listed configurations for all the benchmark experiments listed in Tables~\ref{tab:maintable_cdfsod}-\ref{tab:maintable_rf100} in the main paper without any dataset-specific tunings. 
\paragraph{Computational resources.} Our CD-FSOD~\cite{fu2024cross} experiments are conducted with the MMGDINO-B model, and require 2 RTX 4090/3090 GPUs, or around 48GB GPU memory using other types of GPUs. For the ODinW-13 and RF100-VL experiments, the base model is MMGDINO-L, which is much bigger and requires 4 RTX 4090/3090 GPUs, or around 96GB GPU memory using other types of GPUs. The computational cost can be reduced to half if the batch size is set as 2 instead of 4 when the computational resources are limited.

\section{{Additional ablation study}}
The additional ablation experiments are based on CD-FSOD benchmark~\cite{fu2024cross} with the pre-trained MMGDINO-B model.
\subsection{Parallel decoder layers initialization}
We conducted an ablation experiment on the initialization of the parallel decoder layer in HED to check if the pre-trained weights were still important for the parallel decoder layers.

The results are shown in Table~\ref{tab:importance_hed_pretrained}. We observed that although the single-member in HED has fewer parameters than the conventional detection decoder and is fine-tuned on a small training set (which may make it more prone to overfitting), initializing this part with pre-trained weights remains crucial. Furthermore, if more parameters are randomly initialized (e.g., all parameters of the model), we found that the model becomes difficult to converge, i.e., transfer learning becomes impossible.
\begin{table}
\centering
\scalebox{0.85}{
\begin{tabular}{lccc} 
\toprule
\textbf{CD-FSOD} | \textbf{CD-Mixed} & \textbf{1-shot} & \textbf{5-shot} & \textbf{10-shot} \\ 
\toprule
Randomly Initialized & 33.7 | 25.04 & 44.1 | 36.08 & 47.2 | 38.02 \\
Pre-trained & \textbf{34.9} | \textbf{25.68} & \textbf{45.0} | \textbf{37.14} & \textbf{47.9} | \textbf{40.34} \\
\bottomrule
\end{tabular}}
\caption{{Fine-tuning results with and without randomly initialized parallel layers for HED.}}
\label{tab:importance_hed_pretrained}
\end{table}

\subsection{Learning rate scheduler}
As proposed in the main paper, we argue that the plateau scheduler can better autonomously adjust the learning rate for fine-tuning on larger benchmarks (e.g., RF100-VL ~\cite{robicheaux2025roboflow100vl}), while multi-stage training can also be triggered according to the plateau. However, the cosine scheduler is more commonly used in object detection tasks and can also be applied with HED in our pipeline. Therefore, we conducted ablation experiments using a cosine scheduler. Specifically, we used a cosine scheduler instead of a plateau scheduler when using HED, employing both single-stage and progressive fine-tuning. When using the cosine scheduler, we enabled two-stage training at half of the training epochs.

The results are provided in Table~\ref{tab:ablation_scheduler}. We observed that, when using the cosine scheduler, progressive fine-tuning is still more effective than single-stage. Secondly, the performance of the cosine scheduler is comparable to that of the plateau scheduler.
Intuitively, we believe that since models are more prone to overfitting when fine-tuning with few samples, it is crucial to decrease the learning rate at the right time. Therefore, both schedulers are more suitable than the fixed conventional milestone scheduler. Furthermore, our goal is to find a more general pipeline, in which case the plateau scheduler is a better fit.

\begin{table}
\centering
\scalebox{0.83}{
\begin{tabular}{lccc} 
\toprule
\textbf{CD-FSOD} | \textbf{CD-Mixed} & \textbf{1-shot} & \textbf{5-shot} & \textbf{10-shot} \\ 
\toprule
1-stage (cosine) + HED & 34.6 | 24.92 & 43.6 | 33.52 & 47.2 | 38.30 \\
2-stage (cosine) + HED & 34.6 | 25.46 & 44.2 | 35.34 & 47.4 | 38.42 \\
2-stage (plateau) + HED (Ours) & \textbf{34.9} | \textbf{25.68} & \textbf{45.0} | \textbf{37.14} & \textbf{47.9} | \textbf{40.34} \\
\bottomrule
\end{tabular}}
\caption{{Fine-tuning results for the model using cosine and plateau scheduler with and without our progressive fine-tuning strategies.}}
\label{tab:ablation_scheduler}
\end{table}

\section{Full numerical results}
\subsection{Full results on ODinW-13}
We present the full results on the ODinW-13 benchmark~\cite{zhang2022glipv2} in Tables~\ref{tab:supp_odinw_results_1} to~\ref{tab:supp_odinw_results_4}, which correspond to the average results across three official seeds for the 1-shot, 3-shot, 5-shot, and 10-shot settings, respectively.

We observe that other open-source models achieve comparable zero-shot performance to our base model. However, by applying our proposed training strategies, our model consistently and significantly outperforms the other fine-tuned open-source models.

Furthermore, while GroundingDINO 1.5 Pro~\cite{ren2024groundingpro} is an updated version of GroundingDINO~\cite{liu2024grounding}, featuring a much larger-scale pre-training dataset and a more powerful architecture, which obtained much better zero-shot performance than our base model, our methodology still succeeds in outperforming the fine-tuned version of this strong closed-source model. This result further demonstrates the effectiveness of our proposed method.

\begin{table}[h]
\centering
\begin{tabular}{lc}
\hline
\textbf{Datasets} & \textbf{1-shot} \\
\hline
AerialMaritimeDrone & 26.267 \\
Aquarium & 47.500 \\
CottontailRabbits & 76.967 \\
EgoHands & 67.133 \\
NorthAmericaMushrooms & 81.067 \\
Packages & 65.833 \\
PascalVOC & 68.400 \\
Raccoon & 70.600 \\
ShellfishOpenImages & 66.533 \\
VehiclesOpenImages & 69.667 \\
pistols & 73.267 \\
pothole & 29.233 \\
thermalDogsAndPeople & 78.467 \\
\hline
\textbf{Average} & \textbf{63.1} \\
\hline
\end{tabular}
\caption{\textbf{Average results of ODinW-13 of the three official seeds of runs} for 1-shot setting.}
\label{tab:supp_odinw_results_1}
\end{table}

\begin{table}[h]
\centering
\begin{tabular}{lc}
\hline
\textbf{Datasets} & \textbf{3-shot} \\
\hline
AerialMaritimeDrone & 34.033 \\
Aquarium & 52.767 \\
CottontailRabbits & 77.067 \\
EgoHands & 69.033 \\
NorthAmericaMushrooms & 83.567 \\
Packages & 67.733 \\
PascalVOC & 70.633 \\
Raccoon & 71.167 \\
ShellfishOpenImages & 68.133 \\
VehiclesOpenImages & 69.367 \\
pistols & 71.933 \\
pothole & 40.067 \\
thermalDogsAndPeople & 79.367 \\
\hline
\textbf{Average} & \textbf{65.8} \\
\hline
\end{tabular}
\caption{\textbf{Average results of ODinW-13 of the three official seeds of runs} for 3-shot setting.}
\label{tab:supp_odinw_results_2}
\end{table}

\begin{table}[h]
\centering
\begin{tabular}{lc}
\hline
\textbf{Datasets} & \textbf{5-shot} \\
\hline
AerialMaritimeDrone & 38.200 \\
Aquarium & 53.733 \\
CottontailRabbits & 76.533 \\
EgoHands & 72.767 \\
NorthAmericaMushrooms & 87.000 \\
Packages & 70.833 \\
PascalVOC & 69.933 \\
Raccoon & 75.733 \\
ShellfishOpenImages & 67.733 \\
VehiclesOpenImages & 72.033 \\
pistols & 70.267 \\
pothole & 42.133 \\
thermalDogsAndPeople & 81.367 \\
\hline
\textbf{Average} & \textbf{67.6} \\
\hline
\end{tabular}
\caption{\textbf{Average results of ODinW-13 of the three official seeds of runs} for 5-shot setting.}
\label{tab:supp_odinw_results_3}
\end{table}

\begin{table}[h]
\centering
\begin{tabular}{lc}
\hline
\textbf{Datasets} & \textbf{10-shot} \\
\hline
AerialMaritimeDrone & 39.367 \\
Aquarium & 55.167 \\
CottontailRabbits & 75.200 \\
EgoHands & 71.900 \\
NorthAmericaMushrooms & 88.600 \\
Packages & 73.033 \\
PascalVOC & 70.933 \\
Raccoon & 75.800 \\
ShellfishOpenImages & 67.700 \\
VehiclesOpenImages & 72.000 \\
pistols & 71.633 \\
pothole & 46.333 \\
thermalDogsAndPeople & 83.600 \\
\hline
\textbf{Average} & \textbf{68.6} \\
\hline
\end{tabular}
\caption{\textbf{Average results of ODinW-13 of the three official seeds of runs} for 10-shot setting.}
\label{tab:supp_odinw_results_4}
\end{table}

\subsection{Full results on RF100-VL}
The full fine-tuning results of our method on RF100-VL~\cite{robicheaux2025roboflow100vl} are listed in Table~\ref{tab:aerial_performance} to~\ref{tab:other_performance}. Based on the results of SAM3~\cite{sam3_2025} in the main paper, we argue that for few-shot object detection, powerful visual detection or segmentation foundation models have an advantage in the domains with more natural images, such as Flora-Fauna and even Aerial images. However, in specific domains, such as medical images and documentation content, the results of few-shot fine-tuning are not ideal. We believe this is partly due to the fact that the data distribution of these domains differs from the data during large-scale pre-training. For example, the training data for open-source MM-GroundingDINO~\cite{zhao2024open} is mainly natural images from OpenImage~\cite{OpenImages}, COCO~\cite{lin2014microsoft}, and so on. In this case, even if the zero-shot performance of these specific domains is at the same low level as the zero-shot performance with more natural images, after fine-tuning, the model still needs more data than natural images to achieve a higher detection accuracy in this specific domain.
\begin{table}[h!]
\centering
\begin{tabular}{lc} 
\hline
\textbf{Aerial Datasets} & \textbf{10-shot} \\ 
\hline
aerial-airport & 48.1 \\
aerial-cows & 34.4 \\
aerial-sheep & 48.2 \\
\begin{tabular}[c]{@{}l@{}}apoce-aerial-photographs\\-for-object-detection-\\of-construction-equipment\end{tabular} & 46.2 \\
electric-pylon-detection-in-rsi & 21.1 \\
floating-waste & 36.5 \\
human-detection-in-floods & 36.2 \\
sssod & 53.6 \\
uavdet-small & 39.0 \\
wildfire-smoke & 49.9 \\
zebrasatasturias & 42.9 \\ 
\hline
\textbf{Average} & \textbf{41.5} \\
\hline
\end{tabular}
\caption{\textbf{Results on RF100-VL - Aerial category} for 10-shot Performance (mAP)}
\label{tab:aerial_performance}
\end{table}

\begin{table}[h!]
\centering
\begin{tabular}{lc}
\hline
\textbf{Document Datasets} & \textbf{10-shot} \\
\hline
activity-diagrams & 30.4 \\
all-elements & 39.5 \\
circuit-voltages & 34.5 \\
invoice-processing & 23.7 \\
label-printing-defect-version-2 & 54.1 \\
macro-segmentation & 34.1 \\
paper-parts & 38.8 \\
signatures & 68.9 \\
speech-bubbles-detection & 51.5 \\
wine-labels & 13.6 \\
\hline
\textbf{Average} & \textbf{38.9} \\
\hline
\end{tabular}
\caption{\textbf{Results on RF100-VL - Document category} for 10-shot Performance (mAP)}
\label{tab:document_performance}
\end{table}

\begin{table}[h!]
\centering
\begin{tabular}{lc}
\hline
\textbf{Flora-Fauna Datasets} & \textbf{10-shot} \\
\hline
aquarium-combined & 52.6 \\
bees & 30.9 \\
deepfruits & 64.9 \\
exploratorium-daphnia & 26.7 \\
grapes-5 & 39.3 \\
grass-weeds & 46.7 \\
gwhd2021 & 25.3 \\
into-the-vale & 64.4 \\
jellyfish & 31.8 \\
marine-sharks & 30.2 \\
orgharvest & 11.3 \\
peixos-fish & 30.1 \\
penguin-finder-seg & 73.4 \\
pig-detection & 47.0 \\
roboflow-trained-dataset & 58.1 \\
sea-cucumbers-new-tiles & 59.8 \\
thermal-cheetah & 74.0 \\
tomatoes-2 & 81.8 \\
trail-camera & 72.5 \\
underwater-objects & 12.4\\
varroa-mites-detection--test-set & 20.6 \\
wb-prova & 47.9 \\
weeds4 & 53.0 \\
\hline
\textbf{Average} & \textbf{45.9} \\
\hline
\end{tabular}
\caption{\textbf{Results on RF100-VL - Flora-Fauna category} for 10-shot Performance (mAP)}
\label{tab:flora_fauna_performance}
\end{table}

\begin{table}[h!]
\centering
\begin{tabular}{lc}
\hline
\textbf{Industrial Datasets} & \textbf{10-shot} \\
\hline
-grccs & 50.2 \\
13-lkc01 & 32.6 \\
2024-frc & 62.0 \\
aircraft-turnaround-dataset & 35.4 \\
asphaltdistressdetection & 21.2 \\
cable-damage & 24.2 \\
conveyor-t-shirts & 38.2 \\
dataconvert & 64.8 \\
deeppcb & 46.0 \\
defect-detection & 52.5 \\
fruitjes & 57.1 \\
infraredimageofpowerequipment & 47.0 \\
ism-band-packet-detection & 60.2 \\
l10ul502 & 48.5 \\
needle-base-tip-min-max & 27.9 \\
recode-waste & 39.0 \\
screwdetectclassification & 50.2 \\
smd-components & 59.0 \\
truck-movement & 63.6 \\
tube & 67.8 \\
water-meter & 48.3 \\
wheel-defect-detection & 32.2 \\
\hline
\textbf{Average} & \textbf{46.7} \\
\hline
\end{tabular}
\caption{\textbf{Results on RF100-VL - Industrial category} for 10-shot Performance (mAP)}
\label{tab:industrial_performance}
\end{table}

\begin{table}[h!]
\centering
\begin{tabular}{lc}
\hline
\textbf{Medical Datasets} & \textbf{10-shot} \\
\hline
canalstenosis & 46.6 \\
crystal-clean-brain-tumors-mri-dataset & 56.1 \\
dentalai & 20.4 \\
inbreast & 34.2 \\
liver-disease & 19.7 \\
nih-xray & 8.8 \\
spinefrxnormalvindr & 16.6\\
stomata-cells & 17.8 \\
train & 7.5 \\
ufba-425 & 26.6 \\
urine-analysis1 & 26.0 \\
x-ray-id & 46.4 \\
xray & 8.1 \\
\hline
\textbf{Average} & \textbf{25.8} \\
\hline
\end{tabular}
\caption{\textbf{Results on RF100-VL - Medical category} for 10-shot Performance (mAP)}
\label{tab:medical_performance}
\end{table}

\begin{table}[h!]
\centering
\begin{tabular}{lc}
\hline
\textbf{Other Datasets} & \textbf{10-shot} \\
\hline
buoy-onboarding & 26.1 \\
car-logo-detection & 73.9 \\
clashroyalechardetector & 40.6 \\
cod-mw-warzone & 46.4 \\
countingpills & 85.8 \\
everdaynew & 50.0\\
flir-camera-objects & 32.2 \\
halo-infinite-angel-videogame & 64.4 \\
mahjong & 52.1 \\
new-defects-in-wood & 32.5 \\
orionproducts & 41.9 \\
pill & 43.0 \\
soda-bottles & 28.8 \\
taco-trash-annotations-in-context & 36.8 \\
the-dreidel-project & 53.9 \\
\hline
\textbf{Average} & \textbf{47.2} \\
\hline
\end{tabular}
\caption{\textbf{Results on RF100-VL - Other category} for 10-shot Performance (mAP)}
\label{tab:other_performance}
\end{table}

\subsection{Full results on CD-Mixed set}
We provide numerical results on the proposed CD-Mixed set in Table~\ref{tab:supp_cdmix}, which is consistent with the results shown on Figure 3 in the main paper. The prediction average results are not consistent with the ones in the main paper, since we omitted the FISH dataset in the CD-Mixed set, as the FISH dataset has the same domain (underwater imagery) as the UODD dataset, and this might cause the overlapping object categories.

We first observe that introducing a large number of out-of-distribution (OOD) samples significantly reduces mAP, meaning that some OOD samples contain high-confidence detection results. However, since the samples in our chosen dataset have no overlap in terms of targets (e.g., underwater creatures will not appear in images of industrial defects, clip art will not appear in the images of underwater, and close-ups of insects will not appear in aerial photographs), these high-confidence samples are false positives, meaning results obtained by the model are due to overconfidence. This conclusion has also been found in COCO-FP~\cite{liu2024coco} and aligns with the problem faced by modern deep neural networks~\cite{guo2017calibration}.

Our second observation is that our proposed strategy achieves the best or the second-best performance with and without OOD samples on all few-shot settings, and the performance reduction is also the lowest compared to the other training configurations. Meanwhile, layer-parallelization on the decoder alone did not significantly alleviate the mAP decline problem. We believe that the diversity brought by structural parallelization and the different initialization of each decoder layer gradually weakens with training. However, the input diversity brought by the random initialization of the denoising query can alleviate this problem, allowing the parallel decoder layers to learn different weights, making the final aggregated result more robust and calibrated. Finally, we argue that the design of HED, which not only contains the structural parallelization but also introduces the input randomness, can improve the overall prediction accuracy and provide less overconfident results, i.e., more reliable predictions under few-shot settings.

\begin{table*}[t]
\centering
\scalebox{0.9}{
\begin{tabular}{lcccccc} 
\toprule
\multirow{2}{*}{\begin{tabular}[c]{@{}l@{}}\textbf{Baseline + Progressive ft.}\\\textbf{+ HED w/ re-init Dn Query}\end{tabular}} & \multicolumn{2}{c}{1-shot} & \multicolumn{2}{c}{5-shot} & \multicolumn{2}{c}{10-shot} \\ 
\cmidrule{2-7}
 & w/o OOD & w/ OOD & w/o OOD & w/ OOD & w/o OOD & w/ OOD \\ 
\midrule
ArTaxOr & 49.1 & 44 & 76.8 & 73.2 & 79.2 & 75.7 \\
DIOR & 24.6 & 23.6 & 35.3 & 33.8 & 41.5 & 40.3 \\
NEU-DET & 15.5 & 5.3 & 25.2 & 14.2 & 28.4 & 15.7 \\
UODD & 22.1 & 13.3 & 27.5 & 17.0 & 32.3 & 25.7 \\
clipart1k & 55.6 & 42.2 & 59.4 & 47.5 & 59.6 & 44.3 \\
\rowcolor[rgb]{0.753,0.753,0.753} Avg & 33.38 & \textbf{25.68} & 44.84 & \textbf{37.14} & \textbf{48.2} & \textbf{40.34} \\
\rowcolor[rgb]{0.753,0.753,0.753} Reduction rate  &  & \textbf{-23.07 \%} &  & \textbf{-17.17 \%} &  & \textbf{-16.31 \%} \\ 
\midrule
\multirow{2}{*}{\textbf{Baseline + HED w/ re-init Dn Query}} & \multicolumn{2}{c}{1-shot} & \multicolumn{2}{c}{5-shot} & \multicolumn{2}{c}{10-shot} \\ 
\cmidrule{2-7}
 & w/o OOD & w/ OOD & w/o OOD & w/ OOD & w/o OOD & w/ OOD \\ 
\midrule
ArTaxOr & 40.5 & 32.2 & 71.3 & 67.5 & 78.9 & 74.0 \\
DIOR & 22.4 & 21.0 & 34.5 & 32.9 & 39.3 & 37.9 \\
NEU-DET & 16.4 & 6.6 & 24.2 & 13.0 & 26.1 & 13.3 \\
UODD & 21.7 & 14.2 & 26.3 & 20.2 & 32.1 & 26.1 \\
clipart1k & 55.8 & 40.8 & 60.0 & 45.1 & 60.5 & 42.4 \\
\rowcolor[rgb]{0.753,0.753,0.753} Avg & 31.36 & 22.96 & 43.26 & 35.74 & 47.38 & 38.74 \\
\rowcolor[rgb]{0.753,0.753,0.753} Reduction rate  &  & -26.79 \%&  & -17.38 \%&  & -18.24 \%\\ 
\midrule
\multirow{2}{*}{\begin{tabular}[c]{@{}l@{}}\textbf{Baseline + Progressive ft.}\\\textbf{+ HED w/o re-init Dn Query}\end{tabular}} & \multicolumn{2}{c}{1-shot} & \multicolumn{2}{c}{5-shot} & \multicolumn{2}{c}{10-shot} \\ 
\cmidrule{2-7}
 & w/o OOD & w/ OOD & w/o OOD & w/ OOD & w/o OOD & w/ OOD \\ 
\midrule
ArTaxOr & 49.5 & 42.6 & 74.7 & 70.3 & 79.3 & 74.0 \\
DIOR & 23.9 & 21.8 & 34.5 & 31.7 & 40.1 & 36.0 \\
NEU-DET & 14.5 & 5.1 & 25.5 & 12.0 & 26.6 & 11.0 \\
UODD & 22.7 & 11.4 & 26.9 & 17.1 & 31.0 & 21.8 \\
clipart1k & 57.0 & 40.5 & 60.3 & 44.0 & 61.1 & 45.0 \\
\rowcolor[rgb]{0.753,0.753,0.753} Avg & \textbf{33.52} & 24.28 & 44.38 & 35.02 & 47.62 & 37.56 \\
\rowcolor[rgb]{0.753,0.753,0.753} Reduction rate  &  & -27.57 \%&  & -21.09 \%&  & -21.13 \%\\ 
\midrule
\multirow{2}{*}{\textbf{Baseline + Progressive ft.}} & \multicolumn{2}{c}{1-shot} & \multicolumn{2}{c}{5-shot} & \multicolumn{2}{c}{10-shot} \\ 
\cmidrule{2-7}
 & w/o OOD & w/ OOD & w/o OOD & w/ OOD & w/o OOD & w/ OOD \\ 
\midrule
ArTaxOr & 44.3 & 38.8 & 76.9 & 73.3 & 79.1 & 73.3 \\
DIOR & 23.1 & 21.9 & 35.3 & 33.7 & 42.9 & 41.0 \\
NEU-DET & 14.9 & 3.6 & 24.7 & 7.8 & 26.5 & 12.0 \\
UODD & 22 & 14.2 & 29.1 & 17.7 & 30.8 & 25.0 \\
clipart1k & 56.9 & 40.3 & 60.3 & 41.9 & 60.4 & 40.9 \\
\rowcolor[rgb]{0.753,0.753,0.753} Avg & 32.24 & 23.76 & \textbf{45.26} & 34.88 & 47.94 & 38.44 \\
\rowcolor[rgb]{0.753,0.753,0.753} Reduction rate  &  & -26.3 \%&  & -22.93 \%&  & -19.82 \%\\ 
\midrule
\multirow{2}{*}{\textbf{Baseline}} & \multicolumn{2}{c}{1-shot} & \multicolumn{2}{c}{5-shot} & \multicolumn{2}{c}{10-shot} \\ 
\cmidrule{2-7}
 & w/o OOD & w/ OOD & w/o OOD & w/ OOD & w/o OOD & w/ OOD \\ 
\midrule
ArTaxOr & 38.7 & 31.4 & 72.8 & 66.4 & 78.0 & 74.5 \\
DIOR & 20.5 & 19.0 & 35.1 & 33.9 & 41.6 & 39.7 \\
NEU-DET & 11.6 & 1.3 & 23.1 & 10.6 & 24.9 & 11.5 \\
UODD & 18.9 & 13.7 & 26.0 & 20.0 & 30.9 & 24.3 \\
clipart1k & 57.5 & 38.5 & 60.5 & 47.4 & 61.5 & 40.2 \\
\rowcolor[rgb]{0.753,0.753,0.753} Avg & 29.44 & 20.78 & 43.5 & 35.66 & 47.38 & 38.04 \\
\rowcolor[rgb]{0.753,0.753,0.753} Reduction rate  &  & -29.42 \%&  & -18.02 \%&  & -19.71 \% \\
\bottomrule
\end{tabular}
}
\caption{\textbf{Full results on the prediction performance (mAP) on the clean target test set and the proposed CD-Mixed test set, with the performance reduction in percentage.} The best results are highlighted in bold.}
\label{tab:supp_cdmix}
\end{table*}

\end{document}